\documentclass[10pt,twocolumn,letterpaper]{article}

\usepackage{cvpr}
\usepackage{times}
\usepackage{epsfig}
\usepackage{graphicx}
\usepackage{amsmath}
\usepackage{amssymb}
\usepackage{booktabs}
\usepackage{float}
\usepackage{stfloats}
\usepackage{enumitem}
\usepackage{caption} 
\usepackage[pagebackref=true,breaklinks=true,letterpaper=true,colorlinks,bookmarks=false]{hyperref}

\cvprfinalcopy 

\def\cvprPaperID{8661} 

\ifcvprfinal\fi

\usepackage{aux/custom}

\makeatletter
\newcommand{\printfnsymbol}[1]{%
  \textsuperscript{\@fnsymbol{#1}}%
}
\makeatother

\begin{document}

\title{Spatially Attentive Output Layer for Image Classification}

\author{Ildoo Kim\thanks{Contributed equally.}\qquad Woonhyuk Baek\printfnsymbol{1} \qquad Sungwoong Kim\\
Kakao Brain\\
Seongnam, South Korea \\
{\tt\small \{ildoo.kim, wbaek, swkim\}@kakaobrain.com}
}

\maketitle

\begin{abstract}

Most convolutional neural networks (CNNs) for image classification use a global average pooling (GAP) followed by a fully-connected (FC) layer for output logits. However, this spatial aggregation procedure inherently restricts the utilization of location-specific information at the output layer, although this spatial information can be beneficial for classification. In this paper, we propose a novel spatial output layer on top of the existing convolutional feature maps to explicitly exploit the location-specific output information. In specific, given the spatial feature maps, we replace the previous GAP-FC layer with a spatially attentive output layer (SAOL) by employing a attention mask on spatial logits. The proposed location-specific attention selectively aggregates spatial logits within a target region, which leads to not only the performance improvement but also spatially interpretable outputs. Moreover, the proposed SAOL also permits to fully exploit location-specific self-supervision as well as self-distillation to enhance the generalization ability during training. The proposed SAOL with self-supervision and self-distillation can be easily plugged into existing CNNs. Experimental results on various classification tasks with representative architectures show consistent performance improvements by SAOL at almost the same computational cost.
\end{abstract}


\section{Introduction}
\label{sec:intro}

\begin{figure}
     \centering
     \includegraphics[scale=0.51]{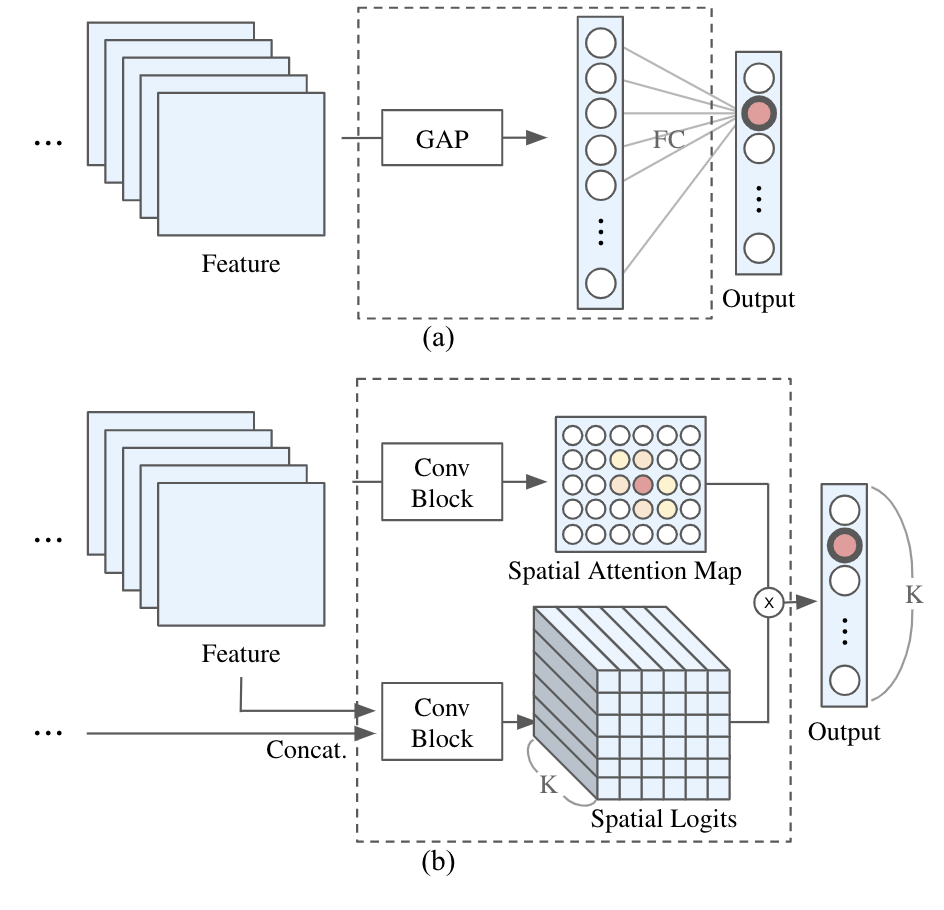}

     \caption{Comparison between (a) the conventional GAP-FC based output layer and (b) the proposed output layer, SAOL. SAOL separately obtains \textit{Spatial Attention Map} and \textit{Spatial Logits} (classification outputs for each spatial location). Then, \textit{Spatial Logits} are weighted averaged by the \textit{Spatial Attention Map} for the final output.}
     \label{fig:overview-layer}
 \end{figure}

Deep convolutional neural networks (CNNs) have made great progress in various computer vision tasks including image classification \cite{krizhevsky2012imagenet, resnet_cvpr}, object detection \cite{RCNN, FasterRCNN, feature-pyramid-net}, and semantic segmentation \cite{FCN, DeepLab}. In particular, there have been lots of researches on modifying convolutional blocks and their connections such as depthwise separable convolution \cite{chollet2017xception}, deformable ConvNet \cite{dai2017deformable}, ResNet \cite{resnet_cvpr}, and NASNet \cite{zoph2018learning} to improve feature representations. However, in contrast to well-developed convolutional architectures for (multi-scale) spatial feature extraction, the output module to generate the classification logits from the feature maps has been almost unchanged from a standard module that is composed of a global average pooling (GAP) layer and fully-connected (FC) layers. Even though it has shown that CNNs with this feature aggregation can retain its localization ability to some extent \cite{NIN, zoph2015iclr, CAM}, in principle, these CNNs have a restriction in full exploitation of benefits from an explicit localization of output logits for image classification.

Recently, the use of localized class-specific responses has drawn increasing attention for image classification, which allows taking the following three main advantages: (1) it can help to interpret the decision making of a CNN through visual explanation \cite{CAM, selvaraju2017gradcam, gradcam++}; (2) a spatial attention mechanism can be used for performance improvement by focusing only on the regions that are semantically relevant to the considered labels \cite{saumya2018iclr, cbam, fei2017cvpr, abn}; and (3) it enables to make use of auxiliary self-supervised losses or tasks based on spatial transformations, which leads to enhanced generalization ability \cite{tellme, gidaris2018unsupervised-rotation, xiaolin2018cvpr, hao2019cvpr, tao2019, lezi2019iccv}.

However, most of the previous methods have obtained spatial logits or attention maps via conventional class activation mapping techniques such as class activation mapping (CAM) \cite{CAM} and gradient-weighted class activation mapping (Grad-CAM) \cite{selvaraju2017gradcam}. They have still utilized the GAP for image-level prediction and thus only located a small part of a target object \cite{tellme} or attended inseparable regions across classes \cite{lezi2019iccv}. While this inaccurate attention mapping hinders its use to improve the classification accuracy, it also has limited an application of self-supervision concerning spatial labeling to maintaining attention consistency under simple spatial transformations such as rotation and flipping \cite{hao2019cvpr} or naive attention cropping and dropping \cite{tao2019}.

Accordingly, we propose to produce explicit and more precise spatial logits and attention maps as well as to apply useful self-supervision by employing a new output module, called {\it Spatially Attentive Output Layer} (SAOL). In specific, from the feature maps, we separately obtain the spatial logits (location-specific class responses) and the spatial attention map. Then, the attention weights are used for a weighted sum of the spatial logits to produce the classification result. Figure \ref{fig:overview-layer} shows an overall structure of the proposed output layer in comparison to the conventional one.

The proposed output process can be considered as a weighted average pooling over the spatial logits to focus selectively on the target class region. For more accurate spatial logits, we aggregate multi-scale spatial logits inspired by decoder modules used for semantic segmentation \cite{FCN, ronneberger2015unet, chen2018encoder}. Note that SAOL can generate spatially interpretable attention outputs directly and target object locations during forward propagation without any post-processing. Besides, the computational cost and the number of parameters of the proposed SAOL are almost the same as the previous GAP-FC based output layer.

Furthermore, we apply two novel location-specific self-supervised losses based on CutMix \cite{yun2019cutmix} to improve the generalization ability. We remark that different from CutMix, which mixes the ground truth image labels proportionally to the area of the combined input patches, the proposed self-supervision utilizes cut and paste of the self-annotated spatial labels according to the mixed inputs. The proposed losses make our spatial logits and attention map more complete and accurate. We also explore a self-distillation by attaching the conventional GAP-FC as well as SAOL and distilling SAOL logits to GAP-FC. This technique can improve performances of the exiting CNNs without changing their architectures at test time.

We conduct extensive experiments on CIFAR-10/100 \cite{cifar} and ImageNet \cite{imagenet} classification tasks with various state-of-the-art CNNs and observe that the proposed SAOL with self-supervision and self-distillation consistently improves the performances as well as generates more accurate localization results of the target objects.


Our main contributions can be summarized as follows:
\begin{itemize}[topsep=2pt]
\setlength\itemsep{-0.3em}
\item The SAOL on top of the existing CNNs is newly proposed to improve image classification performances through spatial attention mechanism on the explicit location-specific class responses.
\item In SAOL, the normalized spatial attention map is separately obtained to perform a weighted average aggregation over the elaborated spatial logits, which makes it possible to produce interpretable attention outputs and object localization results by forward propagation.
\item Novel location-specific self-supervised losses and a self-distillation loss are applied to enhance the generalization ability for SAOL in image-level supervised learning.
\item On both of image classification tasks and Weakly Supervised Object Localization (WSOL) tasks with various benchmark datasets and network architectures, the proposed SAOL with self-supervision consistently improves the performances. Additionally, ablation experiments show the benefits from the more accurate spatial attention as well as the more sophisticated location-specific self-supervision.
\end{itemize}

\section{Related Work}
\label{sec:related}

\begin{figure*}
 \centering
 \includegraphics[scale=0.31]{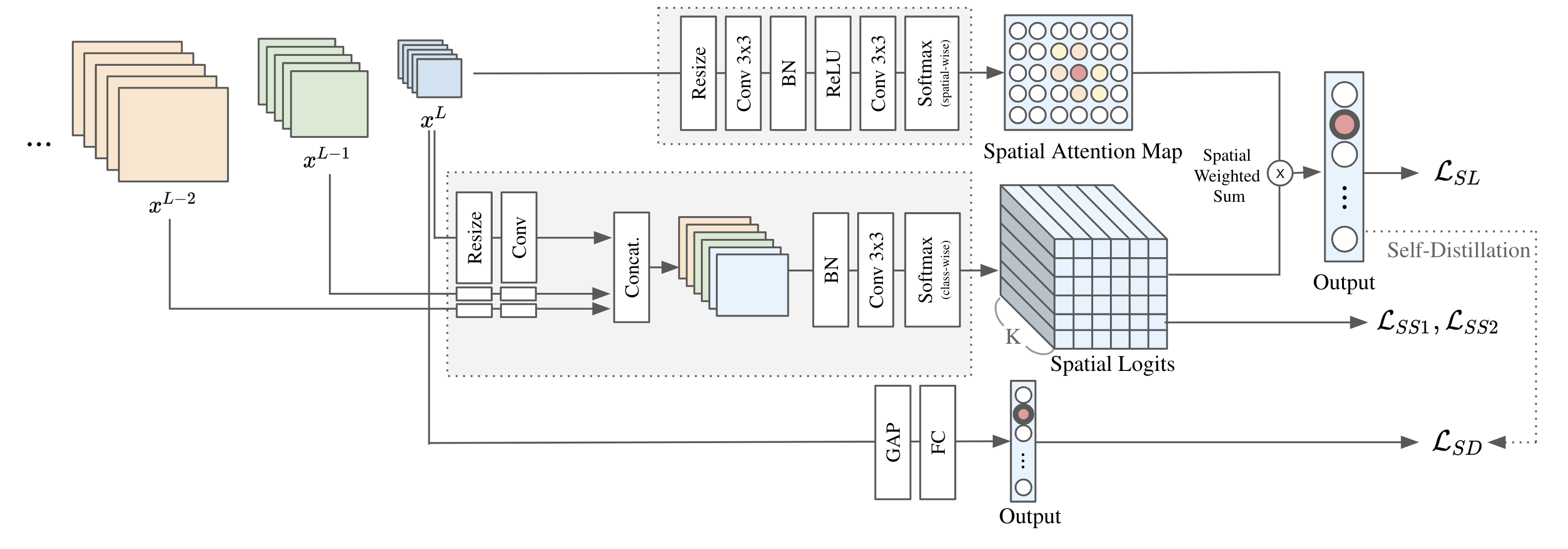}

 \caption{The detailed structure of the proposed SAOL. It produces the spatial attention map and spatial logits, separately. Note that we use additional self-annotated spatial labels to leverage our architecture further. We can also train the conventional GAP-FC based output layer jointly, using self-distillation.}
 \label{fig:specific-layer}
\end{figure*}

{\noindent \bf Class activation mapping.} Class activation mapping methods have been popularly used (1) for visualizing spatial class activations to interpret decision making of the final classification output, (2) for incorporating an auxiliary regularization based on it to boost classification performances, or (3) for performing WSOL. Specifically, CAM \cite{CAM} can obtain an activation map for each class by linearly combining the last convolutional feature maps with the weights associated with that class at the last FC layer. However, CAM needs to replace the FC layer with convolution and GAP to produce the final classification output. On the other hand, Guided Back-propagation \cite{jost2015iclr}, Deconvolution \cite{zeiler2014eccv}, and Grad-CAM \cite{selvaraju2017gradcam} was proposed for generating class-wise attention maps by using gradients in back-propagation without requiring architectural changes. Grad-CAM++ \cite{gradcam++} modified Grad-CAM to localize multiple instances of the same class more accurately using higher-order derivatives. These methods still adapted the GAP for image-level prediction, which often leads to highlighting only on a discriminative but uncompleted part of a target object.

{\noindent \bf  Attention mechanism.}  Several works have been recently explored the use of attention mechanism for image classification and WSOL \cite{saumya2018iclr, cbam, fei2017cvpr, abn}. Residual Attention Network \cite{fei2017cvpr} modified ResNet \cite{resnet_cvpr} by stacking multiple soft attention modules that gradually refine the feature maps. Jetley \etal \cite{saumya2018iclr} proposed a trainable module for generating attention weights to focus on different feature regions relevant to the task of classification at hand. Woo \etal \cite{cbam} introduced a convolutional block attention module that sequentially applies channel and spatial attention modules to refine intermediate feature maps. Attention Branch Network (ABN) \cite{abn} designed a separate attention branch based on CAM to generate attention weights and used them to focus on important feature regions. While all of these attention methods refine intermediate feature maps, we apply the attention mechanism on the output layer to directly improve spatial output logits. Girdhar \etal \cite{girdhar2017attentionalpooling} introduced a more closely related method based on spatial attention for pooling spatial logits on action recognition tasks. Still, they used simple linear mappings only from the last feature map.

{\noindent \bf  CutMix and attention-guided self-supervision.} As an efficient and powerful data augmentation method, CutMix \cite{yun2019cutmix} was recently developed, and it significantly outperforms over previous data augmentation methods such as Cutout \cite{devries2017cutout} and Mixup \cite{zhang2017mixup}. Yet, CutMix cannot guarantee that a randomly cropped patch always has a part of the corresponding target object with the same proportion used for label-mixing. Several recent works derived auxiliary self-supervised losses using attention maps. For example, Guo \etal \cite{hao2019cvpr} proposed to enhance attention consistency under simple spatial transformations, and Hu \etal \cite{tao2019} applied the attention cropping and dropping to data augmentation. Li \etal \cite{tellme} proposed guided attention inference networks that explore self-guided supervision to optimize the attention maps. Especially, they applied an attention mining technique with image cropping to make complete maps; However, these maps are obtained based on Grad-CAM. Zhang \etal \cite{xiaolin2018cvpr} introduced adversarial learning to leverage complementary object regions found by CAM to discover entire objects. Wang \etal \cite{lezi2019iccv} presented new learning objectives for enhancing attention separability and attention consistency across layers. Different from these attention-guided self-supervised learning methods, we design a more sophisticated location-specific self-supervision leveraging CutMix.

\section{Methods}
\label{sec: method}

In this section, we describe the proposed output layer architecture named SAOL and location-specific self-supervised losses and self-distillation loss in detail.

\subsection{Spatially Attentive Output Layer}
\label{sec: spatial-layer}

Let ${\bf x}$ and ${\bf y}$ denote an input image and its one-hot encoded ground truth label, respectively. For CNN-based image classification, an input ${\bf X}^0 = {\bf x}$ is first fed into successive $L$ convolution blocks $\{\Theta_\ell(\cdot)\}_{\ell=1}^{L}$, where intermediate feature maps ${\bf X}^{\ell} \in {\mathbb R}^{C_{\ell} \times H_{\ell} \times W_{\ell}}$ at the block $\ell$ is computed by ${\bf X}^\ell = \Theta_\ell({\bf X}^{\ell-1})$. Here, $H_{\ell}$, $W_{\ell}$, and $C_{\ell}$ are the height, width, and number of channels at the $\ell_{th}$ block. Then, the final normalized output logits ${\bf {\hat y}} \in [0,1]^K$, which can be considered as an output probability distribution over $K$ classes, are obtained by an output layer $O(\cdot)$ such that ${\bf {\hat y}} = O({\bf X}^L)$. In specific, the conventional GAP-FC based output layer $O_{\text{\tiny{GAP-FC}}}(\cdot)$ can be formulated as
\begin{align}
    \label{eq:final-gap-fc-output}
   \!\! {\bf {\hat y}} = O_{\text{\tiny{GAP-FC}}}({\bf X}^L) = \mathrm{softmax}\bigg (({\bf {\bar x}}_{\text{\tiny{GAP}}}^{L})^T {\bf W}^{FC} \bigg ),
\end{align}
where ${\bf {\bar x}}_{\text{\tiny{GAP}}}^{L} \in {\mathbb R}^{C_L \times 1}$ denotes the spatially aggregated feature vector by GAP, and ${\bf W}^{FC} \in {\mathbb R}^{C_L \times K}$ is the weight matrix of the output FC layer. Here, $({\bf {\bar x}}_{\text{\tiny{GAP}}}^{L})_c = \frac{\sum_{i,j} ({\bf X}_c^L)_{ij}}{H_{\ell}W_{\ell}}$, where $({\bf X}_c^L)_{ij}$ is the $(i,j)_{th}$ element of the $c_{th}$ feature map ${\bf X}_c^L$ at the last block. Instead of this aggregation on the last feature map, our method produces output logits explicitly on each spatial location and then aggregates them selectively through the spatial attention mechanism.

Specifically, the proposed SAOL, $O_{\text{\tiny{SAOL}}}(\cdot)$, first produces \textit{Spatial Attention Map}, ${\bf A} \in [0,1]^{H_o \times W_o}$, and \textit{Spatial Logits}, ${\bf Y} \in [0,1]^{K \times H_o \times W_o}$, separately. Here, it is noted that we set $H_o = H_L$ and $W_o = W_L$ by default. The attention values are normalized via softmax across the spatial positions while we take softmax on the spatial logits across classes: $\sum_{i,j} {\bf A}_{ij}=1, \forall k$ and $\sum_{k} ({\bf Y}_k)_{ij}=1, \forall i,j$. Then, we generate the final output logits by a spatially weighted sum of the spatial logits as follows:
\begin{align}
    \label{eq:final-saol-output}
    {\bf {\hat y}}_k = O_{\text{\tiny{SAOL, k}}}({\bf X}^L) = \sum_{i,j} {\bf A}_{ij}({\bf Y}_k)_{ij}, ~~\forall k,
\end{align}
where ${\bf {\hat y}}_k$ is the output logit of the $k_{th}$ class. These attention weights indicate the relative importance of each spatial position regarding the classification result. 


The architecture in SAOL is described in detail in Figure \ref{fig:specific-layer}. First, to obtain the spatial attention map ${\bf A}$, we feed the last convolutional feature maps ${\bf X}^L$ into two-layered convolutions followed by the softmax function. At the same time, for the sake of the precise spatial logits, we combine multi-scale spatial logits, motivated by previous decoder modules for semantic segmentation \cite{FCN, ronneberger2015unet, chen2018encoder}. In specific, at each of the selected blocks, the feature maps are mapped to the intermediate spatial logits through convolutions after resized to the output spatial resolution. Then, a set of the intermediate spatial logits are concatenated and re-mapped to the final spatial logits ${\bf Y}$ by another convolution layer and the softmax function. Note that in contrast to CAM \cite{CAM} and Grad-CAM \cite{selvaraju2017gradcam}, this SAOL can directly generate spatially interpretable attention outputs or target object locations using ${\bf A}$ and ${\bf Y}$ in a feed-forward manner. This makes it possible to use location-specific regularizers during training, as presented in the next subsection.


\subsection{Self-Supervised Losses}
\label{sec: self-superv-loss}

\begin{figure}
     \centering
     \includegraphics[scale=0.49]{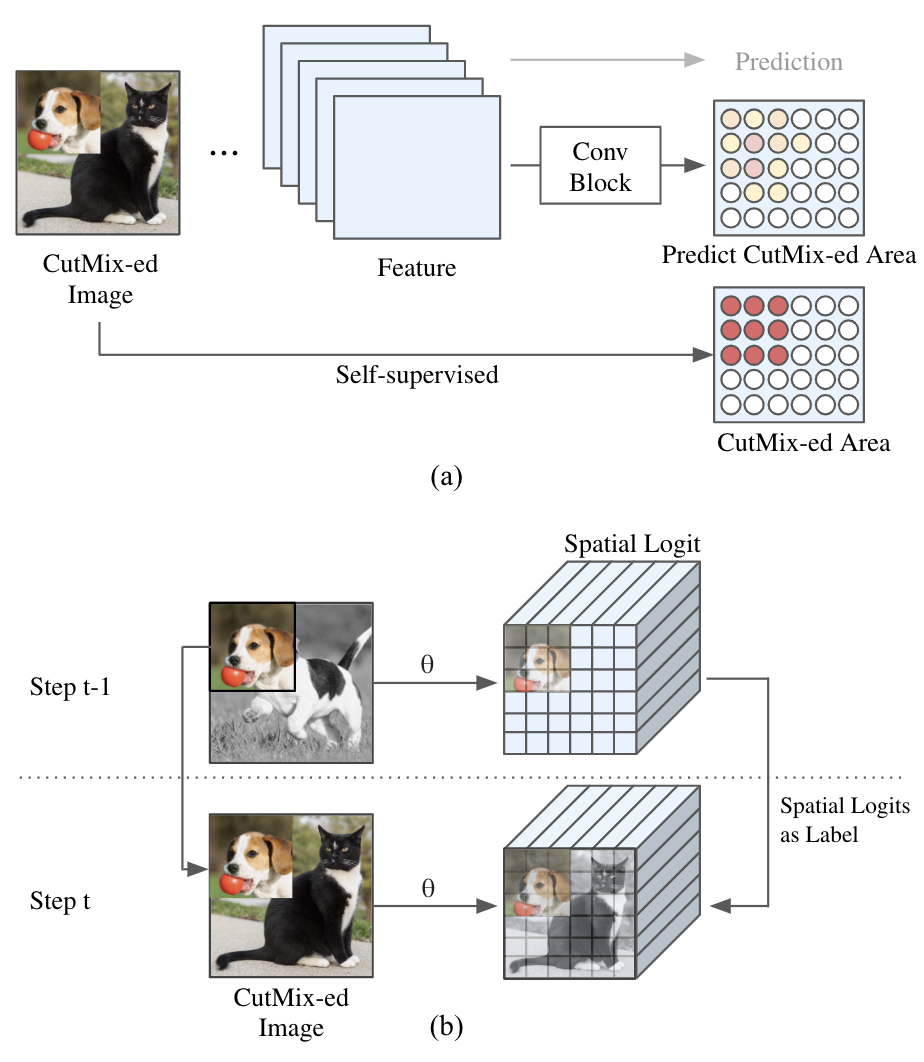}

     \caption{The proposed two self-supervisions based on CutMix for SAOL: (a) ${\mathcal L}_{SS1}$ and (b) ${\mathcal L}_{SS2}$.}
     \label{fig:overview-self-annotated}
 \end{figure}

The proposed SAOL performs well when trained even only with the general cross-entropy loss ${\mathcal L}_{CE}$ as our supervised loss such that $\mathcal{L}_{SL} = \mathcal{L}_{CE}({\bf {\hat y}}_{\text{\tiny{SAOL}}}, {\bf y})$\footnote{We let ${\bf {\hat y}}_{\text{\tiny{GAP-FC}}}$ and ${\bf {\hat y}}_{\text{\tiny{SAOL}}}$ denote the final output logits from the GAP-FC based output layer and those from SAOL, respectively.}. However, in order to fully utilize location-specific output information to boost the classification performance, we add two novel spatial losses inspired by CutMix \cite{yun2019cutmix} and self-supervised learning methods \cite{gidaris2018unsupervised-rotation, lee2019rethinking}. 

CutMix generates a new training sample $({\bf x}', {\bf y}')$ by mixing a certain sample $({\bf x}_B, {\bf y}_B)$ and a random patch extracted from an another sample $({\bf x}_A, {\bf y}_A)$ as follows: 
\begin{equation}
    \label{eq:cutmix}
    \begin{aligned}
    {\bf x}' &= \mathbf{M}\odot {\bf x}_A + (\mathbf{1-M})\odot {\bf x}_B, \\
    {\bf y}' &= \lambda {\bf y}_A + (1-\lambda){\bf y}_B,
    \end{aligned}
\end{equation}
where $ \mathbf{M} $ denotes a binary mask for cropping and pasting a rectangle region, and $ \lambda $ is a combination ratio sampled using the beta distribution. This label-mixing strategy implies that a cut region should have the meaning as much as the size of the cropped area in the context of its label. However, this assumption would often be incorrect since a randomly cropped patch can fail to capture a part of the corresponding target object, especially when the target object is small.

Specifically, we use two additional self-annotated spatial labels and self-supervised losses, as illustrated in Figure \ref{fig:overview-self-annotated}.
Given a CutMix-ed input image, the first self-supervised loss $\mathcal{L}_{SS1}$ uses $\mathbf{M}$ as an additional ground truth label after resizing to ${H_o \times W_o}$. We add an auxiliary layer similar to the attention layer to predict $\hat{\mathbf{M}} \in [0, 1]^{H_o \times W_o}$. Since $\mathbf{M}$ is the binary mask, the binary cross-entropy loss is used as
\begin{align}
    \mathcal{L}_{SS1} = \mathcal{L}_{BCE}(\hat{\mathbf{M}}, \mathbf{M}).
\end{align}
The second self-supervised loss $\mathcal{L}_{SS2}$ we propose is to match the spatial logits in the pasted region of the mixed input with the spatial logits in the cut region of the original data as follows:

\begin{align}
    \mathcal{L}_{SS2} = D_{KL}(\mathbf{M}\odot {\bf Y}', \mathbf{M}\odot {\bf Y}_A),
\end{align}
where $D_{KL}$ represents the Kullback–Leibler divergence\footnote{It is actually the average Kullback–Leibler divergence over spatial positions.}, and ${\bf Y}_A$ denotes the spatial logits of ${\bf x}_A$. Since these self-supervisions regularize the network either to identify the specific pasted location or to produce the same spatial logits in the pasted region, these can lead to spatially consistent feature representations and accordingly, improved performances. Note that we update the network through the gradients only from $\mathbf{M}\odot {\bf Y}'$.

\subsection{Self-Distillation Loss}
\label{sec: self-distil-loss}

Since one can insert the proposed SAOL in the existing CNNs, we utilize both the previous GAP-FC based output layer and SAOL, as shown in Figure \ref{fig:specific-layer}, during training. Specifically, we come up with knowledge transfer from SAOL to the existing output layer. For this, we devise a self-distillation loss $\mathcal{L}_{SD}$ with the two final output logits separately obtained by the two output layers from a given input image, as follows:
\begin{align}
    \label{eq:self-distillation-loss}
    \mathcal{L}_{SD} = D_{KL}({\bf {\hat y}}_{\text{\tiny{SAOL}}}, {\bf {\hat y}}_{\text{\tiny{GAP-FC}}}) + \beta \mathcal{L}_{CE}({\bf {\hat y}}_{\text{\tiny{GAP-FC}}}, {\bf y}),
\end{align}
where $\beta$ is the relative weight between the two loss terms, which was similarly used in other self-distillation methods \cite{zhang2019self-distill, lee2019rethinking}. We set $\beta = 0.5$. At test-time, we take only one of the two output modules to produce the classification result. If we select the GAP-FC based output layer, we can improve the classification performances of the exiting CNNs without computational tax at test time, although it is negligible.

In the end, the final loss ${\mathcal L}$ that we use during training is defined as
\begin{align}
    \label{eq:final-loss}
    \mathcal{L} = \mathcal{L}_{SL} + \mathcal{L}_{SS1} + \mathcal{L}_{SS2} + \mathcal{L}_{SD},
\end{align}
where further improvement may be possible using different ratios of losses. 

\section{Experiments}

We evaluate our SAOL with self-supervision and self-distillation compared to the previous methods. We first study the effects of our proposed method on several classification tasks in Section \ref{sec: exp-classification}. Then, to conduct a quantitative evaluation for the obtained attention map, WSOL experiments were performed in Section \ref{sec: exp-wsol}. 

All experiments were implemented in PyTorch \cite{paszke2017automatic}, by modifying the official CutMix source code\footnote{https://github.com/clovaai/CutMix-PyTorch}. For a fair comparison, we tried not to change the hyper-parameters from baselines such as CutMix \cite{yun2019cutmix} and ABN \cite{abn}. We simultaneously trained both of SAOL and the GAP-FC based output layer via the proposed self-distillation loss in an end-to-end manner. At test time, we obtained the classification results by either SAOL or the GAP-FC based output layer. 



\subsection{Image Classification Tasks}
\label{sec: exp-classification}

\subsubsection{CIFAR-10, CIFAR-100 Classification}

\begin{table*}[t!]\center
\centering
\begin{tabular}{l | c | c | c c}\toprule

&Baseline &\multicolumn{2}{c}{Ours} \\
Model &GAP-FC &SAOL &{\tiny self-distilled} GAP-FC \\
\midrule 

Wide-ResNet 40-2 \cite{wide-resnet} &94.80~ &\textbf{95.33 (\footnotesize{+0.53})} &95.31 (\footnotesize{+0.51}) \\
Wide-ResNet 40-2 + CutMix \cite{yun2019cutmix} &96.11~ &\textbf{96.44 (\footnotesize{+0.33})} &96.44 (\footnotesize{+0.33}) \\
Wide-ResNet 28-10 \cite{wide-resnet} &95.83~ &\textbf{96.44 (\footnotesize{+0.61})} &96.42 (\footnotesize{+0.59}) \\
Wide-ResNet 28-10 + CutMix \cite{yun2019cutmix} &97.08~ &\textbf{97.37 (\footnotesize{+0.29})} &97.36 (\footnotesize{+0.28}) \\
ResNet-110 \cite{resnet_cvpr} &93.57* &\textbf{95.18 (\footnotesize{+1.61})} &95.06 (\footnotesize{+1.49}) \\
ResNet-110 + CutMix \cite{yun2019cutmix} &95.77~ &\textbf{96.21 (\footnotesize{+0.44})} &96.17 (\footnotesize{+0.40}) \\
DenseNet-100 \cite{densenet} &\textbf{95.49*} &95.31 (\footnotesize{-0.18}) &95.35 (\footnotesize{-0.14}) \\
DenseNet-100 + CutMix \cite{yun2019cutmix} &95.83~ &\textbf{96.27 (\footnotesize{+0.44})} &96.19 (\footnotesize{+0.36}) \\
PyramidNet200 + ShakeDrop \cite{shakedrop} &97.13~ &\textbf{97.33 (\footnotesize{+0.20})} &97.31 (\footnotesize{+0.18}) \\
PyramidNet200 + ShakeDrop + CutMix \cite{yun2019cutmix} &97.57~ &\textbf{97.93 (\footnotesize{+0.36})} &97.92 (\footnotesize{+0.35}) \\

\bottomrule
\end{tabular}

\caption{Classification Top-1 accuracies (\%) on CIFAR-10. Results from the original papers are denoted as *.}\label{tab:exp-cifar10}

\vspace*{0.2 cm}

\begin{tabular}{l | c | c | c c}\toprule

&Baseline &\multicolumn{2}{c}{Ours} \\
Model &GAP-FC &SAOL &{\tiny self-distilled} GAP-FC \\
\midrule 

Wide-ResNet 40-2 \cite{wide-resnet} &74.73~ &\textbf{76.50 (\footnotesize{+1.77})} &76.18 (\footnotesize{+1.45}) \\
Wide-ResNet 40-2 + CutMix \cite{yun2019cutmix} &78.21~ &\textbf{79.53 (\footnotesize{+1.32})} &79.04 (\footnotesize{+0.83}) \\
Wide-ResNet 28-10 \cite{wide-resnet} &80.13~ &80.89 (\footnotesize{+0.76}) &\textbf{81.16 (\footnotesize{+1.03})} \\
Wide-ResNet 28-10 + CutMix \cite{yun2019cutmix} &82.41~ &\textbf{83.71 (\footnotesize{+1.30})} &83.71 (\footnotesize{+1.30}) \\
ResNet-110 \cite{resnet_cvpr} &75.86* &77.15 (\footnotesize{+1.29}) &\textbf{77.23 (\footnotesize{+1.37})} \\
ResNet-110 + CutMix \cite{yun2019cutmix} &77.94~ &\textbf{78.02 (\footnotesize{+0.08})} &77.94 (\footnotesize{+0.00}) \\
DenseNet-100 \cite{densenet} &\textbf{77.73*} &76.84 (\footnotesize{-0.89}) &76.25 (\footnotesize{-1.48}) \\
DenseNet-100 + CutMix \cite{yun2019cutmix} &78.55~ &\textbf{79.25 (\footnotesize{+0.70})} &78.90 (\footnotesize{+0.35}) \\
PyramidNet200 + ShakeDrop \cite{shakedrop} &84.43~ &84.72 (\footnotesize{+0.29}) &\textbf{84.95 (\footnotesize{+0.52})} \\
PyramidNet200 + ShakeDrop + CutMix \cite{yun2019cutmix} &86.19~ &86.95 (\footnotesize{+0.76}) &\textbf{87.03 (\footnotesize{+0.84})} \\

\bottomrule
\end{tabular}

\caption{Classification Top-1 accuracies (\%) on CIFAR-100. Results from the original papers are denoted as *.}\label{tab:exp-cifar100}

\end{table*}

The first performance evaluation for image classification is carried out on CIFAR-10 and CIFAR-100 benchmark \cite{cifar}, one of the most extensively studied classification tasks. We used the same hyper-parameters for Wide-ResNet \cite{wide-resnet} from AutoAugment \cite{cubuk2018autoaugment}. ResNet and DenseNet models were trained with the same settings for ABN \cite{abn} to compare each other. For PyramidNet200 (widening factor ${\bar \alpha}=240$), we used the same hyper-parameters used in CutMix \cite{yun2019cutmix}, except for the learning rate and its decay schedule. We used 0.1 as the initial learning rate for cosine annealing schedule \cite{cosine-anneal}. While our baselines did not obtain much better results with this slight change, the proposed SAOL achieved noticeable performance improvements. Every experiment was performed five times to report its average performance. 

Table \ref{tab:exp-cifar10} and Table \ref{tab:exp-cifar100} compare the baseline and the proposed method on CIFAR-10 and CIFAR-100, respectively. The proposed SAOL outperformed the baseline consistently across all models except DenseNet-100. In addition, in most cases for CIFAR-10, SAOL gave clear improvements over self-distilled GAP-FC. However, our self-distilled GAP-FC was also consistently better than the baseline. This means that even without spatial supervision such as object localization label, SAOL can learn spatial attention appropriately and eventually performs better than averaging features. This consistent improvement was also retained when we additionally used CutMix during training. 

We also compare SAOL with recently proposed ABN \cite{abn}. There are similarities between the two methods in respect of using the attention map. However, SAOL uses the attention map to aggregate spatial output logits. In contrast, ABN makes use of the attention mechanism only on the last feature maps and adapts the previous GAP-FC layer. For ResNet-110 and DenseNet-100, we trained models with the same hyper-parameters used in ABN. ResNet-110 and DenseNet-100 with ABN achieved the accuracies of 95.09\%, 95.83\% on CIFAR-10 and 77.19\%, 78.37\% on CIFAR-100, respectively. These results indicate that models with SAOL perform much better than models with ABN. We emphasize that ABN also requires more computations. To be specific, ResNet-110 with ABN requires 5.7 GFLOPs, while ResNet-110 with SAOL only requires 2.1 GFLOPs. As the original ResNet-110 computes as much as 1.7 GFLOPs, not only SAOL is more effective and efficient than ABN, but also it provides a way to keep the amount of computation intact through self-distillation.

\subsubsection{ImageNet Classification}

\begin{table*}[!htp]\centering

\begin{tabular}{l | c | c | c c}\toprule
 & Baseline &\multicolumn{2}{c}{Ours} \\
Model & GAP-FC &SAOL &{\tiny self-distilled} GAP-FC \\\midrule
ResNet-50 \cite{resnet_cvpr} &76.32 / 92.95* &\textbf{77.11 / 93.59} &76.66 / 93.25 \\
ResNet-50 + CutMix \cite{yun2019cutmix} &78.60 / 94.10* &\textbf{78.85 / 94.24} &78.09 / 94.00 \\
ResNet-101 \cite{resnet_cvpr} &78.13 / 93.71* &\textbf{78.59 / 94.25} &78.22 / 93.82 \\
ResNet-101 + CutMix \cite{yun2019cutmix} &79.83 / 94.76* &\textbf{80.49 / 94.96} &80.24 / 94.84 \\
ResNext-101 \cite{resnext} &78.82 / 94.43* &\textbf{79.23 / 95.03} &79.23 / 94.97 \\
ResNext-101 + CutMix \cite{yun2019cutmix} &80.53 / 94.97* &\textbf{81.01 / 95.15} &80.81 / 95.03 \\
ResNet-200 \cite{resnet_cvpr} &78.50 / 94.20~ &\textbf{79.31 / 94.54} &78.92 / 94.37 \\
ResNet-200 + CutMix \cite{yun2019cutmix} &80.70 / 95.20~ &\textbf{80.82 / 95.19} &80.73 / 95.21 \\
\bottomrule

\end{tabular}
\caption{ImageNet classification Top-1 / Top-5 accuracies (\%). Results from the original papers are denoted as *.  }\label{tab:exp-imagenet}

\end{table*}

We also evaluate SAOL on ILSVRC 2012 classification benchmark (ImageNet) \cite{imagenet} which consists of 1.2 million natural images for training and 50,000 images for validation of 1,000 classes. We used the same hyper-parameters with CutMix \cite{yun2019cutmix}. For faster training, we just changed the batch size to 4,096 with a linearly re-scaled learning rate and a gradual warm-up schedule, as mentioned in \cite{facebook1hour}. We also replaced all convolutions in SAOL with depthwise-separable convolutions \cite{howard2017mobilenets} to reduce computations. We found that in many situations, this convolution change made a marginal difference in performances.

\begin{figure}
     \centering
     \includegraphics[scale=0.51]{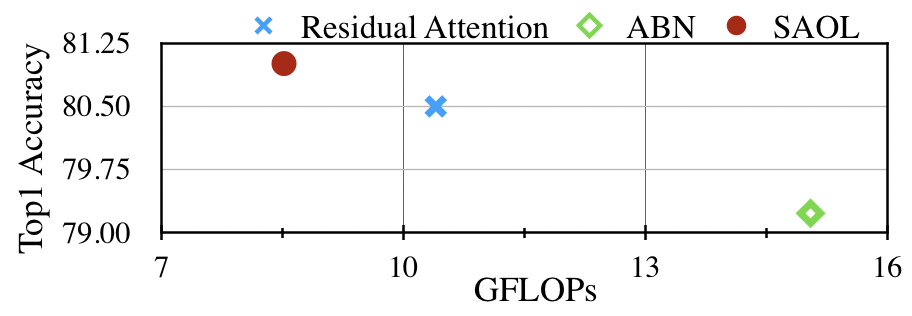}

     \caption{Comparison of different attention models on ImageNet. Attention layers are added on the same ResNet-200 backbone. Our model (SAOL) outperforms previous methods \cite{abn, fei2017cvpr} using negligible computational overhead. } 
     \label{fig:quantitive-analysis}
 \end{figure}

Table \ref{tab:exp-imagenet} shows performances with diverse architectures. We quoted results from the CutMix paper except for ResNet-200, which was not tested by CutMix. We trained all models with the same hyper-parameters for a fair comparison. Our results indicate that models with SAOL outperformed the models with GAP-FC consistently. For example, ResNet-101 architecture trained with CutMix regularization scored 79.83\% of top-1 accuracy, which is improved from 78.13\% without CutMix. For both cases, SAOL further improves the model by 0.46\% and 0.66\% without and with CutMix, respectively. We remark that adding our SAOL requires 6\% more computations only (from 7.8 GFLOPs to 8.3 GFLOPs), which is efficient compared to the previous methods. As shown in Figure \ref{fig:quantitive-analysis}, SAOL performed better than both of Residual Attention Network \cite{fei2017cvpr} and ABN \cite{abn}, especially even with much smaller computational cost.

\subsubsection{Ablation Study}
\label{sec: exp-ablation}

In this section, we conduct ablation experiments for many factors in SAOL to measure their contributions towards our outperforming results. 
\newline

{\noindent \bf Effectiveness of Multi-level Feature Aggregation for Spatial Logits.} SAOL uses features not only from the last convolution block but from multiple intermediate blocks for producing the spatial logits. In detection and segmentation tasks, majority of works \cite{chen2018encoder}\cite{ronneberger2015unet}\cite{DeepLab}\cite{feature-pyramid-net} similarly used multiple feature layers in a decoder to be more size-invariant. We experimented on CIFAR-100 to verify performance changes according to different numbers of features to be combined to generate the spatial logits for SAOL, and Table \ref{tab: ab-deepsupervision} shows the obtained results. 
Performances tend to be improved with more feature layers for spatial logits.

\begin{table}[]\centering

\begin{tabular}{l | c c }\toprule


&WResNet 40-2 &WResNet 28-10 \\\midrule
Conv Block 3 &75.68 &79.99 \\
Conv Block 2+3 &76.18 &80.70 \\
Conv Block 1+2+3 &\textbf{76.50} &\textbf{80.89} \\

\bottomrule
\end{tabular}

\caption{Performance comparisons on CIFAR-100 according to different combinations of feature blocks used for producing the spatial logits. WResNet stands for Wide-ResNet. Wide-ResNet has three convolutional blocks, and we denote the $i$th block as Conv Block $i$. }\label{tab: ab-deepsupervision}

\end{table}
{\noindent \bf Effectiveness of Self-Supervision.} To verify the benefits from the proposed two self-supervised losses, we conducted experiments with Wide-ResNet 40-2 on CIFAR-10 and CIFAR-100 (C-100), and the results are shown in Table \ref{tab: ab-ss}. Similar to the baseline model, SAOL was also improved with the original CutMix regularization alone. However, additional incorporating $\mathcal{L}_{SS1}$ or $\mathcal{L}_{SS2}$ further enhanced the performances. Using both of self-supervised losses with SAOL led to the best performance.

\begin{table}[]
\centering

\begin{tabular}{l | c | c}\toprule
&CIFAR-10 &C-100 \\\midrule

Baseline (GAP-FC) &94.80 &74.73 \\
Baseline + CutMix &96.11 &78.21 \\
Baseline + CutMix + $ \mathcal{L}_{SS1} $ &96.04 &78.14 \\
\hline
SAOL &95.33 &76.50 \\
SAOL + CutMix &96.21 &78.44 \\
SAOL + CutMix + $ \mathcal{L}_{SS1} $ &96.19 &78.92 \\
SAOL + CutMix + $ \mathcal{L}_{SS2} $ &96.30 &78.60 \\
SAOL + CutMix + $ \mathcal{L}_{SS1} $ + $ \mathcal{L}_{SS2} $ &\textbf{96.44} &\textbf{79.53} \\
\bottomrule
\end{tabular}

\caption{Influences of CutMix and its additional self-supervised losses for Wide-ResNet 40-2 on CIFAR-10/100.}\label{tab: ab-ss}

\end{table}

Note that we also tried to use $\mathcal{L}_{SS1}$ on the baseline. For this, we attached an auxiliary layer on the last convolution block to produce a spatial map predicting the CutMix region and trained the original image classification loss and $\mathcal{L}_{SS1}$ jointly. As a result, the use of $\mathcal{L}_{SS1}$ did not improve the performance of the baseline. We conjecture that SAOL worked well with $\mathcal{L}_{SS1}$ since it tried to learn the attention map for classification outputs simultaneously. We leave a more detailed investigation of this for future work.

{\noindent \bf Effectiveness of Self-Distillation.} We also conducted experiments on CIFAR-100 to measure the effectiveness of our self-distillation. Instead of distilling outputs from SAOL, the standard cross-entropy (CE) loss was solely applied to the GAP-FC auxiliary layer during training. The results are shown in Table \ref{tab: ab-sd}. Irrespective of the selected output layer at test time, training both of SAOL and the GAP-FC based output layer with the same CE loss led to performance drop compared to the use of our self-distillation loss $\mathcal{L}_{SD}$, even though it still outperformed the baseline. This indicates that the knowledge transfer from robust SAOL to the conventional GAP-FC based output layer by our self-distillation is beneficial to performance improvement.

\begin{table}[H]\centering

\begin{tabular}{l | c | c | c | cc}\toprule

&\multicolumn{2}{c|}{WResNet 40-2} &\multicolumn{2}{c}{WResNet 28-10}  \\
&SAOL &GAP-FC &SAOL &GAP-FC \\ \midrule

Baseline &N/A &74.73 &N/A &80.13 \\
CE &75.75 &75.28 &80.36 &80.21 \\
$ \mathcal{L}_{SD} $ &\textbf{76.50} &\textbf{76.18} &\textbf{80.89} &\textbf{81.16} \\

\bottomrule

\end{tabular}
\caption{Evaluation on the effectiveness of self-distillation. }\label{tab: ab-sd}

\end{table}


\subsection{Weakly-Supervised Object Localization Task}
\label{sec: exp-wsol}



\begin{table*}[tp]\center

\begin{tabular}{l l r c | c | c}\toprule
Model &Method &GFLOPs &Backprop. &\multicolumn{1}{p{2.1cm}|}{\centering CUB200-2011 \\ Loc Acc (\%)} &\multicolumn{1}{p{1.8cm}}{\centering ImageNet \\ Loc Acc (\%)} \\
\midrule

ResNet-50 \cite{resnet_cvpr}                               &CAM \cite{CAM} &4.09 &O &49.41* &46.30* \\
ResNet-50 \cite{resnet_cvpr} + CutMix \cite{yun2019cutmix} &CAM \cite{CAM} &4.09 &O &54.81* &47.25* \\
ResNet-50 \cite{resnet_cvpr}                               & ABN \cite{abn} &7.62 &X &56.91~ &44.65~ \\
ResNet-50 \cite{resnet_cvpr} + CutMix \cite{yun2019cutmix} & SAOL (Ours) &4.62 &X               &52.39~ &45.01~ \\

\bottomrule

\end{tabular}

\caption{Weakly supervised object localization results on CUB200-2011 test set and ImageNet validation set. The asterisk * indicates that the score is from the original paper.} \label{tab:exp-wsol}

\end{table*}

\begin{figure*}
     \centering
     \includegraphics[scale=0.385]{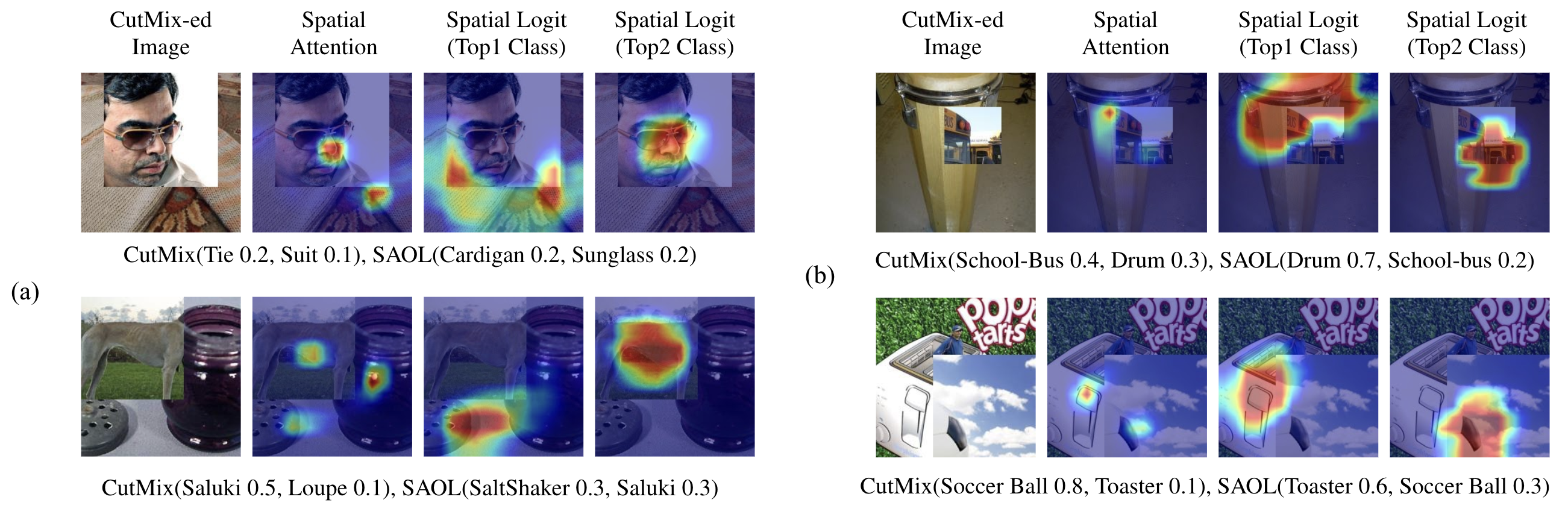}

     \caption{Qualitative analysis of attention maps by SAOL with ResNet-50. From the left: CutMix-ed image, spatial attention map, heatmap of spatial output logit for top-2 classes. (a) Examples that previous CutMix model \cite{yun2019cutmix} failed to correctly predict objects with top-2 classes' scores. (b) Examples that previous CutMix model predicted small objects over-confidently.} 
     \label{fig:qualitative-analysis}
\end{figure*}

To evaluate the spatial attention map by SAOL quantitatively, we performed experiments with ResNet-50 models for the tasks of WSOL. We followed the evaluation strategy of the existing WSOL method \cite{CAM}. A common practice in WSOL is to normalize the score maps using min-max normalization to have a value between 0 and 1. The normalized output score map can be binarized by a threshold, then the largest connected area in the binary mask is chosen. Our model was modified to enlarge the spatial resolutions of the spatial attention map and spatial logits to be $14 \times 14$ from $7\times 7$ and finetuned ImageNet-trained model. The obtained spatial attention map and spatial logits are combined as an elemental-wise product to yield a class-wise spatial attention map.

As the result are shown in Table \ref{tab:exp-wsol}, our method achieves competitive localization accuracy on ImageNet and CUB200-2011 \cite{CUB_200_2011}, compared to previous well-performing methods \cite{choe2019attention-drop-wsol, yun2019cutmix}. It is noticeable that our competitive method requires much fewer computations to generate an attention map for object localization. While it is common to use CAM \cite{CAM}, burdensome backward-pass computations are unavoidable. Recently proposed ABN \cite{abn} can produce an attention map with the single forward pass; however, it modifies the backbone network with a computationally-expensive attention mechanism. SAOL adds much less computational taxes while it performs competitively. We also emphasize that our results were obtained without any sophisticated post-processing, which is required by many WSOL methods. Utilizing sophisticated post-processing as well as training with a larger attention map may improve the result further.

Figure \ref{fig:qualitative-analysis} visualizes the spatial attention map and the spatial logits obtained by SAOL on CutMix-ed image. Our spatial attention map focuses on the regions corresponding to the general concept of objectness. On the other hand, the spatial output logits show class-specific activation maps which have high scores on the respective target object regions. In the situation where two objects are mixed, the attention map by SAOL localizes each object well, and moreover its scores reflect the relative importance of each object more accurately. 

\section{Conclusion}
\label{sec:conclusion}

We propose a new output layer for image classification, named spatially attentive output layer (SAOL). Outputs from the novel two branches, spatial attention map and spatial logits, generate the classification outputs through an attention mechanism. The proposed SAOL improves the performances of representative architectures for various tasks, with almost the same computational cost. Moreover, additional self-supervision losses specifically designed for SAOL also improve the performances further. The attention map and spatial logits produced by SAOL can be used for weakly-supervised object localization (WSOL), and it shows promising results not only for WSOL tasks but also towards interpretable networks. We will continue this research to develop better decoder-like output structures for image classification tasks and to explore a more sophisticated use of self-annotated spatial information without human labor.

\clearpage

{\small
\bibliographystyle{ieee_fullname}
\bibliography{aux/references}
}

\end{document}